\documentclass{article}

\usepackage{amsfonts}
\usepackage{times}
\usepackage{xcolor}
\usepackage{soul}
\usepackage{graphicx}
\usepackage[utf8]{inputenc}
\usepackage[small]{caption}
\usepackage{algorithm}
\usepackage[noend]{algpseudocode}
\usepackage{subcaption}
\usepackage{tabulary} 
\usepackage{colortbl,booktabs}
\usepackage{threeparttable}  
\usepackage{diagbox}
\usepackage{amsmath}
\usepackage{authblk}
\usepackage[utf8]{inputenc}
\usepackage[final]{corl_2018} 

\title{Intervention Aided Reinforcement Learning for Safe and Practical Policy Optimization in Navigation}

\author[1]{Fan Wang}
\author[1]{Bo Zhou}
\author[1]{Ke Chen}
\author[2]{Tingxiang Fan}
\author[1]{Xi Zhang}
\author[1]{Jiangyong Li}
\author[1]{Hao Tian}
\author[2]{Jia Pan}

\affil[1]{Baidu Inc.}
\affil[1]{\{wang.fan, zhoubo01, chenke06, zhangxi13, lijiangyong01, tianhao\}@baidu.com}
\affil[2]{City University of HongKong}
\affil[2]{\{tingxiangfan,panjia1983\}@gmail.com}

\begin{document}
\maketitle
\vspace*{-.30in}


\begin{abstract}
Combining deep neural networks with reinforcement learning has shown great potential in the next-generation intelligent control. However, there are challenges in terms of safety and cost in practical applications. In this paper, we propose the Intervention Aided Reinforcement Learning (IARL) framework, which utilizes human intervened robot-environment interaction to improve the policy. We used the Unmanned Aerial Vehicle (UAV) as the test platform. We built neural networks as our policy to map sensor readings to control signals on the UAV. Our experiment scenarios cover both simulation and reality. We show that our approach substantially reduces the human intervention and improves the performance in autonomous navigation\footnote{A visual demonstration of the learned policy is available at \href{https://youtu.be/jdMntfs9dYQ}{Youtube}; Our code is based on the PaddlePaddle Reinforcement Learning library(PARL), which is released in \href{https://github.com/PaddlePaddle/PARL/tree/develop/examples/IARL/}{Github}}, at the same time it ensures safety and keeps training cost acceptable.
\end{abstract}



\section{Introduction}
	\vspace*{-.05in}
	In contrast to the intense studies of deep Reinforcement Learning(RL) in games and simulations~\cite{duan2016benchmarking}, employing deep RL to real world robots remains challenging, especially in high risk scenarios. Though there has been some progresses in RL based control in realistic robotics~\cite{kober2013reinforcement, levine2016end, rusu2016sim, zhu2017target}, most of those previous works does not specifically deal with the safety concerns in the RL training process. For majority of high risk scenarios in real world, deep RL still suffer from bottlenecks both in cost and safety. As an example, collisions are extremely dangerous for UAV, while RL training requires thousands of times of collisions. Other works contributes to building simulation environments and bridging the gap between reality and simulation~\cite{rusu2016sim, zhu2017target}. However, building such simulation environment is arduous, not to mention that the gap can not be totally made up. 
    

To address the safety issue in real-world RL training, we present the Intervention Aided Reinforcement Learning (IARL) framework. Intervention is commonly used in many automatic control systems in real world for safety insurance. It is also regarded as an important evaluation criteria for autonomous navigation systems, e.g. the disengagement ratio in autonomous driving\footnote{For details, see: \href{https://www.dmv.ca.gov/portal/dmv/detail/vr/autonomous/disengagement_report_2017}{Autonomous Vehicle Disengagement Reports}}. In this work, instead of using disruptive events such as collisions as feedbacks, we try to optimize the policy by avoiding human intervention. Though utilizing human intervention to avoid fatal mistakes in RL training has been applied to Atari games recently~\cite{saunders2018trial}, our work is featured in real world applications. Besides, we propose a more general framework to take human intervention into account: We define the expression of the intervention as a combination of a unknown classifier deciding when to seize control and a reference control policy; We redefine the behavior policy as blending of the policy to be optimized and the intervention; Then we try to reduce the probability of intervention and learn from the reference policy at the same time.
    
    In this work we are mainly interested in navigating an Unmanned Aerial Vehicle (UAV) in cluttered and unstructured environments. The agent is required to approach a specified goal at certain velocity without collision. Compared with traditional navigation, the proposed navigation policy is based on a more end-to-end architecture. The policy model takes multi-modal observations including 3D depth image, supersonic sensors, velocity, direction to the goal etc. as input and performs attitude control. We compared IARL with different learning algorithms including classic imitation learning and reinforcement learning in simulation, and tested IARL in the realistic UAV navigation task. We demonstrate that IARL learn to avoid collision nearly perfectly compared with the other baselines, even though it has never seen a single collision in the training process. In the meantime, IARL substantially reduces the human intervention more than the other methods. Furthermore, our method can be applied to various autonomous control systems in reality such as autonomous driving cars.



\section{Related Works}


	
    There has been abundant works on visual navigation system\cite{bonin2008visual}, among which our work is closely related to map-less reactive visual navigation systems~\cite{oleynikova2015reactive,syre2017reactive}. Traditional reactive controllers either rely on artificially designed rules, such as boundary following, which highly adapts to specific obstacles~\cite{matveev2013problem,oleynikova2015reactive}, or use empirical models such as Artificial Potential Fields (APF~\cite{lacroix1998reactive,vadakkepat2000evolutionary}), which builds virtual potential fields for collision avoidance. The above works requires artificial tuning of a lot of hyper-parameters and hand-written rules. Some of those methods show competitive performance in specific scenarios, but they are not easy to be extended to unstructured scenarios.
    
	Imitation Learning takes an important role in intelligent controllers in recent years, where policy is learned by reducing the difference between itself and a reference policy(typically the human demonstration). Among them, the supervised learning (or behavior cloning) generates the training trajectories directly from the reference policy\cite{liu2017learning,bojarski2016end}. It is relatively simple, but the performance is usually unstable due to the inconsistency between the training and testing trajectories. Ross et al. proposed the Data Aggregation(DAgger~\cite{ross2011reduction}) as an improvement to supervised learning, by generating trajectories with the current policy. Their subsequent work~\cite{ross2013learning} successfully employed DAgger to collision-free navigation on an UAV. Yet, DAgger remains expensive, and it requires a tricky labeling process that is hard to operate in practice. 
    
    Our work is also related to recent progresses in reinforcement learning, especially the deterministic policy gradient dealing with continuous control~\cite{lillicrap2015continuous, schulman2015trust, schulman2017proximal}. We use Proximal Policy Optimization(PPO)~\cite{schulman2017proximal} with General Advantage Estimation~\cite{schulman2015high}(GAE) as baseline, which has been widely accepted for its robustness and effectiveness. However, most RL algorithms can not be directly deployed to the high risk scenarios due to the safety issue, thus we made the comparison in simulation only. In addition, we also use the trick of combining imitation learning and reinforcement learning to accelerate the training process, which is closely related to the previous work of \cite{rajeswaran2017learning,levine2013guided}. 


\section{Algorithms}
	\subsection{Backgrounds}
	In this part we introduce the notations as well as the representations of different benchmarking algorithms. A Markov Decision Process(MDP) is described by $(\mathcal{S}, \mathcal{A}, \mathcal{R}, \mathcal{P})$, where $\mathcal{S}$ and $\mathcal{A}$ represent the state and action space, $\mathcal{R}$ represents the rewards, which is typically the real number field, and $\mathcal{P}$ represents the transition probability. We use the notation $t\in[0,T]$ to represent the time step, where $T$ is the length of an episode. $s_t \in \mathcal{S}$, $a_t \in \mathcal{A}$ and $r_t \in \mathcal{R}$ denote the specific state, action and reward at step $t$. We use the notation $\mu$ to represent the \emph{deterministic} policy that we want to optimize. We have $a_t = \mu_{\theta}(s_t)$, with $\theta$ representing the trainable parameters to be optimized. A more general \emph{stochastic} policy is defined as a distribution over $\mathcal{S} \times \mathcal{A}$. Usually a Gaussian distribution $\mathcal{N}$ is used for stochastic policy in continuous action space, which can be written as $\pi_{\theta}(a|s_t) = \mathcal{N}(\mu_{\theta}(s_t), \sigma)$. $\sigma$ can either be a hyper-parameter or be optimized all-together with the other trainable parameters.

	\textbf{Imitation Learning}.
 	The behavior cloning is a straightforward approach in imitation learning. If we use the mean square error for the target function(loss function), it can be written as
   	\begin{equation}
	\label{Eq:Supervised Learning}
	L^{SL}(\theta) \propto \mathbb{E}_{\pi^{ref}}[\sum_{t \leq T} (\mu_{\theta}(s_t) - a^{ref}_t)^2],
	\vspace*{-.1in}
    \end{equation}
where $a^{ref}_t \in \pi^{ref}$ denotes the reference action, the notation $\mathbb{E}_{\pi}$ represent the expectation over trajectories that follow policy $\pi$(the behavior policy). Usually the data collection and the policy updating are separate and independent in such supervised learning process, which leads to inconsistency between training and testing. The DAgger proposed in \cite{ross2011reduction} substitutes the behavior policy $\pi^{ref}$ with $\pi_{\theta}$ to alleviate the discrepancy. We omit the loss function, as it is similar to Equation.~\ref{Eq:Supervised Learning}.
    
	\textbf{Policy Gradient}.
	We use the similar formulations of policy gradients as Schulman et al.~\cite{schulman2015high}, where the target function $L_{PG}$ is written as: 
	\begin{equation}
	\label{Eq:Policy Gradient}
	L^{PG}(\theta) \propto -\mathbb{E}_{\pi_{\theta}}[\sum_{t \leq T} \Psi_t \log{\pi_{\theta}(a_t | s_t)}].
	\end{equation}
	There are different expressions for $\Psi_t$ in Equation~\ref{Eq:Policy Gradient}. $\Psi_t = \sum_{0 \leq \tau \leq T} r_{\tau}$ gives the REINFORCE~\cite{williams1992simple}. The Actor-Critic method~\cite{sutton1998reinforcement} uses one-step temporal difference error that is given by $\Psi_t^{(1)} = r_t + \gamma V^{\pi}_{\phi}(s_{t+1}) - V^{\pi}_{\phi}(s_t)$, where $V^{\pi}_{\phi}$ represents the discounted value function on policy $\pi$, and $\gamma$ denotes the discount factor. The GAE(\cite{schulman2015high} is represented by $\Psi_t^{GAE} \propto \sum_{t \leq \tau < T} (\gamma \lambda)^{\tau-t} \Psi_{\tau}^{(1)}$. 

	The calculation of $\Psi_t$ mentioned above requires on-policy generation of trajectories. In case external intervention is introduced, the policy optimization in Equation~\ref{Eq:Policy Gradient} can not be directly applied. In the next subsection we reformulate the policy optimization paradigm with intervention.

\subsection{Intervention Aided Reinforcement Learning}

	We assume that the expert in the IARL system is represented by $(\pi^{ref}, \mathcal{G})$, where $\mathcal{G}(s)\in[0,1]$ is the possibility of intervening the system when the expert observes the agent being at state $s$. Notice that both $\pi^{ref}$ and $\mathcal{G}$ is decided by the expert himself, which is not known in advance to the agent. The expert samples a decision $g_t\in\{0,1\}$ according to $\mathcal{G}(s_t)$ at time $t$, determining whether the agent is intervened eventually. We write the actual action as $a^{mix}_t = g_t \cdot a^{ref}_t + (1 - g_t) \cdot a_t$ with $g_t \sim Bernoulli(\mathcal{G}(s_t))$. $a^{mix}_t$ is regarded to be sampled from the mixed policy of the experts and the agent, which is denoted as $\pi^{mix}_{\theta}$ and given by:
	\begin{equation}
	\label{Eq:IARL-Action}
 	\pi^{mix}_{\theta}(a_t|s_t) = \mathcal{G}(s_t) \pi^{ref}(a_t|s_t) + (1 - \mathcal{G}(s_t))\pi_{\theta}(a_t|s_t).
	\end{equation}
	Note that the behavior policy (the policy under which the data is collected) in IARL is $\pi^{mix}_{\theta}$, not $\pi_\theta$. In order to approximate $\Psi_t$ on-policy, we optimize $\theta$ through the mixed policy $\pi^{mix}_{\theta}$ instead of $\pi_{\theta}$.

	In order to prevent intervention, we also reshape the reward function in IARL. We treat intervention as failures, and reshape the reward by punishing the intervention $g_t$ with
	\begin{equation}
	\label{Eq:IARL-Reward}
 	r^{mix}_t = r_t - g_t \cdot b,
	\end{equation}
	where $b$ is a predefined hyper-parameter. Notice that all the collisions have been prevented by the expert intervention, thus the agent learns to avoid collisions only from avoiding the intervention. By replacing $r_t$ with $r^{mix}_t$ and $\pi_\theta$ with $\pi^{mix}_{\theta}$, the derivative of policy loss in Equation~\ref{Eq:Policy Gradient} to $\theta$ is written as Equation~\ref{Eq:IARL-Policy-Initial}. Note that any derivation of the reference policy $\pi^{ref}$ vanishes, and we use $K(s, a)$ to replace $(1 - \mathcal{G}(s))\frac{\pi_{\theta}(a | s)}{\pi^{mix}_{\theta}(a | s)}$. 
	\begin{equation}
	\label{Eq:IARL-Policy-Initial}
	\begin{split}
 	&\frac{\partial L^{PG}(\theta)}{\partial (\theta)} \propto
 		- \mathbb{E}_{\pi^{mix}_{\theta}}[\sum_{t \leq T} \Psi_t \frac{K(s_t, a^{mix}_t)}{\pi_{\theta}(a^{mix}_t | s_t)} \frac{\partial \pi_{\theta}(a^{mix}_t | s_t)}{\partial (\theta)}]
	\end{split}
	\end{equation}
	 We can see that $K(s, a) \in [0, 1]$ and $K(s, a) = 0$ if $\mathcal{G}(s) = 1$, and $K(s, a)=1$ if $\mathcal{G}(s) = 0$. $K(s, a)$ can be interpreted as a ``mask" that blocks the gradient to $\pi_\theta$ when $\mathcal{G}(s) = 1$. For those states that $\mathcal{G}(s) = 1$, the action is intervened by $\pi^{ref}$ and the trajectory is not determined by $\pi_\theta$, thus the gradient vanishes. However, $\mathcal{G}(s_t)$ as well as $\pi^{mix}_{\theta}$ cannot be directly approximated, and thus $K(s_t,a^{mix}_t)$ is not known exactly. We turn to use a simple approximation of $K(s_t, a^{mix}_t) \approx (1-g_t)$. The error of the approximation is small when $g(s_t)$ is close to $0$ or $1$; it indeed may result in large errors in other cases, but these errors turns out to be acceptable in our experiments. We rewrite the loss of IARL as:
	\begin{equation}
	\label{Eq:IARL-Policy}
	\begin{split}
 	&L^{\text{IARL}}(\theta) \propto
 		- \mathbb{E}_{\pi^{mix}_{\theta}}[\sum_{g_t=0} \Psi_t \log{\pi_{\theta}(s_t, a^{mix}_t)}],
	\end{split}
	\end{equation}
	In Equation~\ref{Eq:IARL-Policy} the summation is running over steps satisfying $g_t=0$ only, while the steps with $g_t=1$ are completely ``masked''. Notice that the value function still backup through all states to give $\Psi_t$, thus Equation~\ref{Eq:IARL-Policy} tends to avoid the intervention beforehand. However, Equation~\ref{Eq:IARL-Policy} does not improve the performance of the policy $\pi_\theta$ once the intervention has taken place (e.g., when $\mathcal{G}(s_t)$ is close to 1). This is harmful to the robustness of policy $\pi_\theta$, as $\pi_\theta$ is likely to malfunction in the states of ``intervened zone''. To alleviate the defects, we introduce an additional imitation loss into Equation~\ref{Eq:IARL-Policy}. The imitation loss matches the optimizing policy $\pi_\theta$ to the reference policy $\pi^{ref}$, which leads to:
	\begin{equation}
	\label{Eq:IARL-Imitation-Policy}
	\begin{split}
 	&L^{\text{IARL-IL}}(\theta) \propto
 		- \mathbb{E}_{\pi^{mix}_{\theta}}[\sum_{g_t=0} \Psi_t \log{\pi_{\theta}(a^{mix}_t | s_t)} + \alpha\sum_{g_t=1}||\mu_{\theta}(s_t) - a^{ref}_t||^2].
	\end{split}
	\end{equation}
	An intuitive explanation of Equation~\ref{Eq:IARL-Imitation-Policy} is that the first term in the right side avoids intervention beforehand, and the second term imitates the reference policy $\pi^{ref}$ in the ``intervened zone''. The hyper-parameter $\alpha$ balances the importance of the two parts. The training process of IARL is presented in Algorithm~\ref{Algorithm:IARL}. It can be easily extended to PPO, we leave those details in the supplements. We call Equation~\ref{Eq:IARL-Policy} IARL (No Imitation) and Equation~\ref{Eq:IARL-Imitation-Policy} IARL (Imitation).

	\begin{algorithm}
	\caption{Intervention Aided Reinforcement Learning}
	\label{Algorithm:IARL}
	\begin{algorithmic}[1]
	\State Initialize $\phi$ for value function $V_\phi$, $\theta$ for policy $\pi_\theta$ and $\pi^{mix}_\theta$.
	\For{iterations = 1,2,3,... until convergence}
	\For{episodes = 1,2,3,... until enough data collected}
	\For{t = 1,2,3,... until finished}
	\State Observe $s_t$, receive $g_t$ from the expert
    \If{$g_t = 1$} receive $a^{ref}_t$ from the expert, $a^{mix}_t = a^{ref}_t$
    \Else{} $a^{mix}_t \sim \pi_{\theta}$
    \EndIf
	\State Take action $a^{mix}_t$
	\State Receive $r_t$, push the record $(s_t, a^{mix}_t, r_t, g_t)$ to memory
	\EndFor
	\EndFor
	\State Optimize $\theta$ by minimizing $L^{IARL}$ for several epochs
	\State Update $V_{\phi}$ using Temporal Difference or Monte Carlo estimate
	\EndFor
	\end{algorithmic}
	\end{algorithm}
    

\section{System Setup}
    In order to validate IARL, we build the UAVs and the control systems in both simulation and reality. The configurations in both environments are similar to each other. In this section we briefly introduce the architecture of the control system and the neural network structure of the decision model(the policy).
    
\subsection{Control Architecture}
	The overall control system includes the decision model to be optimized, two assistant Proportional-Integral-Derivative (PID) controllers and an underlying double closed loop PID controller. The input to our system includes image flows captured by binocular stereo vision, supersonic sensors in its three directions (left, right and front) and Inertial Measurement Unit(IMU).
    
\begin{figure}
	\centering
  	\includegraphics[width=0.75\linewidth]{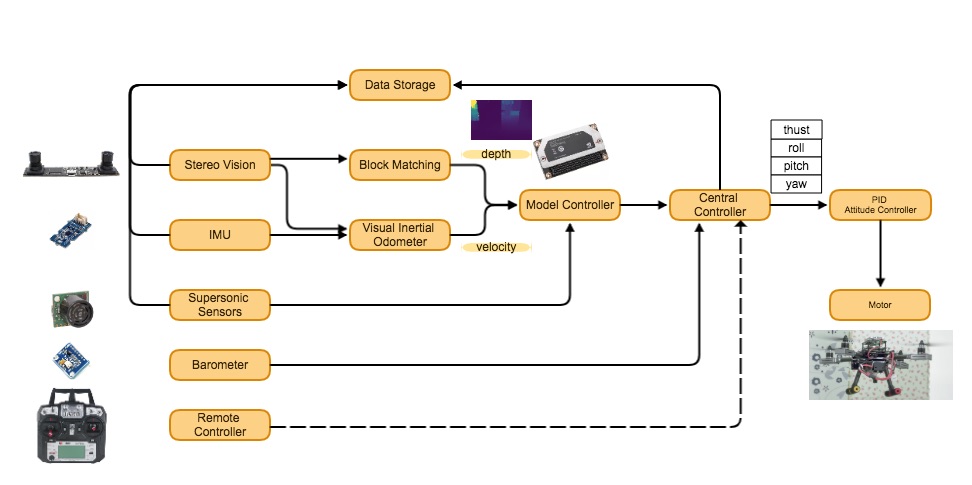}
  	\caption{Diagram of the architecture of the controllers on the drone.}
  	\label{fig:Framework}
    \vspace*{-.2in}
	\end{figure}

    The command of pitch and roll angle of the UAV is decided by the policy model. We use two additional controller to deal with the yaw and the thrust, which keeps the drone at a fixed height and heading angle. The central controller sends the full attitude commands to the underlying control module, which controls to the attenuators of the motor. The remote controller is able to override those commands at any time if a switch is turned on by the human expert, which corresponds to the $g$ in Algorithm~\ref{Algorithm:IARL}. A sketch of the controller architecture can be found in Figure~\ref{fig:Framework}. The output of the policy model is calculated at a frequency of 10Hz, and the underlying controllers exports at the frequency of 400Hz. 

\subsection{Model Structure}
	Our model does not use raw signals as the input directly. Instead, we introduce several preprocessing steps. The IMU and stereo cameras are combined and preprocessed to generate velocities and positions as well as the depth information. We use an open-source SDK \footnote{The boteye SDK: https://github.com/baidu/boteye}. We also calculate the velocity and the direction to the goal from raw signals and use them as the observation, details of the features used can be found in the supplements.

\begin{figure}
  	\centering
  	\includegraphics[width=0.75\linewidth]{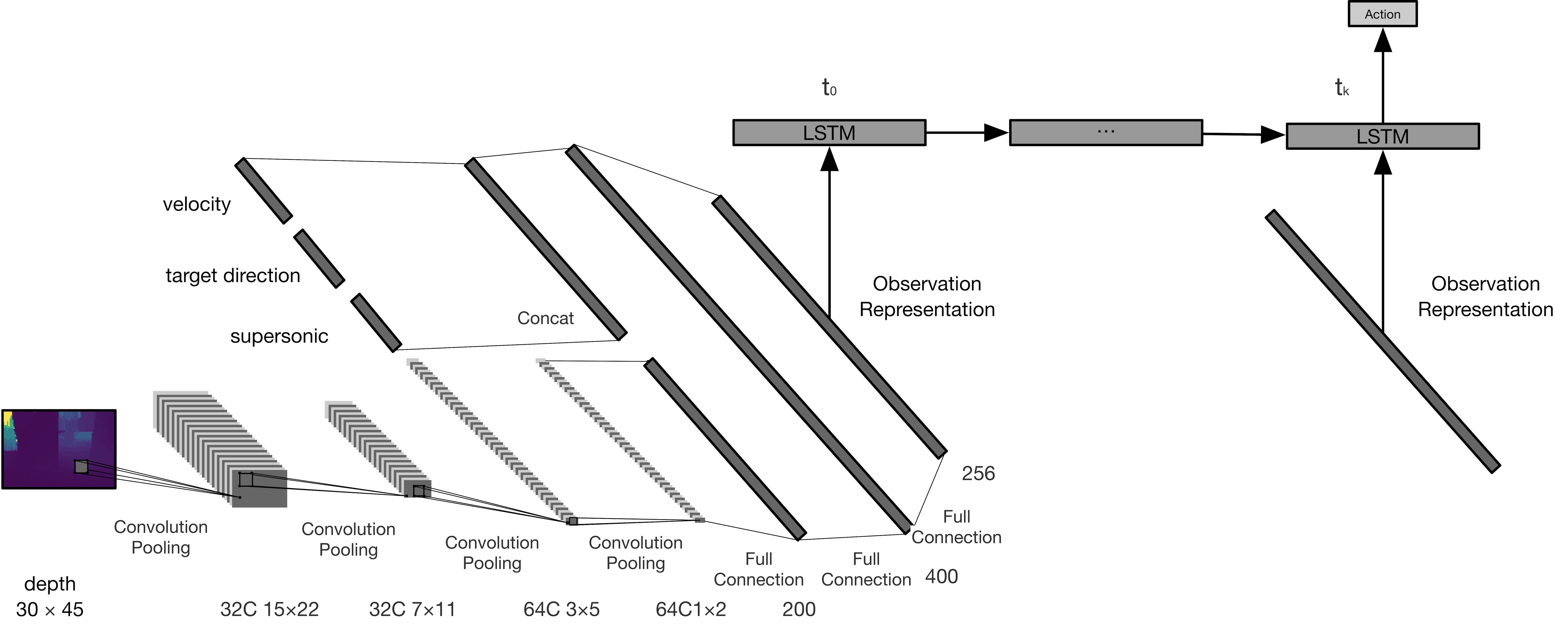}
	\caption{A sketch of the policy model structure.}
    \label{fig:ModelStructure}
    \vspace*{-.2in}
	\end{figure}

	The policy network $\pi_{\theta}$ and value network $V_{\phi}$ share the same model structure but have different output and independent parameters. For each frame of depth image we apply four convolution-pooling layers. The hidden representation is then concatenated with the other features. Traditional reactive controller such as APF typically uses only observation of one single frame. However, the real system is more close to a Partially Observable MDP due to the noisy sensor signals. It is found that multiple frames of observations are helpful to reducing the noise and making up the missing information in each single observation. We build a model that is similar to the architecture reported by Hausknecht et al.~\cite{hausknecht2015deep}. The LSTM unit~\cite{hochreiter1997long} is applied to encode the observations from last 2.5 seconds in order. We apply a tanh layer to restrict the output to $[-1, +1]$, which corresponds to the pitch/roll angle of $\pm 0.1 rad$. The structure of the model is shown in Figure~\ref{fig:ModelStructure}. We also investigated the performance of two other different models. We used a fixed dataset collected from the simulation using reference policy, which contains 1 million frames of training data and 50,000 frames of testing data. The mean square error(MSE) performance of different structures are compared: The single observation (2D convolution + MLP) gives MSE = $0.02+$; Stacking multiple frames (3D convolution + MLP) gives MSE = $0.0066$; Recurrent networks (2D convolution + LSTM) gives MSE = $0.0044$. The test result demonstrates the superiority of the proposed model structure.

\section{Experiments}

\subsection{Simulation Environments}

	\textbf{Simulator.}  We use the Robot Operation System (ROS) and Gazebo 7~\cite{koenig2004design} for simulation. The drone is simulated using the open-source hector-quad-rotor package on ROS~\cite{2012simpar_meyer} . Some of the kinetic hyper-parameters are re-adjusted to match our drone in reality. An Intel Realsense sensor is attached to the simulated drone to represent the stereo vision. In the simulation, the velocity and position are directly acquired from the simulator instead of being calculated from the raw observation.
	
    \textbf{Environments.} The simulation region is $10m \times 10m$ in scale. The drone is required to start from the start point of ($x=5$m, $y=-1$m) and find its way to the goal on top of the region ($y = 11$m). The height and yaw angle of the UAV is kept unchanged. We manually designed three kinds of obstacles in Gazebo 7, including $1m \times 3m$ walls, $3m \times 3m$ walls, and buckets of $0.7m$ diameter and $1.5m$ height. We randomly scattered a certain amount of obstacles in the region, the scenarios without a valid inner path between the starting point and the goal is dropped. 100 different scenarios were finally generated with different random seeds. 90 of them were treated as training scenes, while the left 10 were the testing scenes.
	
    \textbf{Expert simulation.} Training with real human operators in simulation is a relatively expensive choice. In order to further reduce the cost, we build a simulated expert, which gives $\mathcal{G}$ and $\pi^{ref}$ automatically. We use ground truth information provided by Gazebo 7 to simulate the expert, which is hidden from the agents. The details of experts simulation are left in the supplements. The simulated expert help to avoid around $95\%$ collision during IARL training. In order to make a fair comparison, we want the IARL agent to \textbf{only} learn from intervention. Thus we dropped the left $5\%$ collided episodes in the IARL training process.

	\textbf{Reward.} The basic reward function for RL and IARL is set to $r = \max(5 + 20v_d - 50v^2, 0.0)$, where $v_d$ is the projection of the velocity in the direction of the goal. The term of $20v_d$ encourages the drone toward the desired goal. The term $50v^2$ is to slow down the drone to avoid risky accelerations. The maximum reward shall be achieved if the drone moves at the speed of $0.20$m/s towards the goal. The collision with any obstacle triggers the failure of an episode, which terminates the episode and gives additional $-25$ punishment. If no collision is detected, the episode terminates when arriving the goal or reaching at most $600$ steps($60$ seconds). In IARL, we use $b=5$ for the intervention punishment such that $r^{mix} = r - 5 g_t$. 
    
	\textbf{Evaluation.}  We perform two kinds of tests for each comparing method and each test scenario: The \emph{Normal Test} evaluates $\pi_{\theta}$ by removing the external interference $\pi^{ref}$, $R_{\text{NT}} = \sum r_t$ is used as the criteria. The \emph{Intervention Aided Test} keeps the external interference $\pi^{ref}$. $R_{\text{IAT}} = \sum r^{mix}_t$ is used as the criteria. Another criteria in \emph{Intervention Aided Test} is called the intervention rate (\textbf{IR}), defined by $\textbf{IR} = (\sum_t g_t) / (\sum_t 1)$. This evaluation metric is similar to the disengagement rate evaluation in autonomous vehicle. All our evaluation results are averaged among the $10$ test scenarios.

\subsection{Methods for comparison} 

	\textbf{SL}. We collect 1 million frames under policy $\pi^{ref}$ in 90 training scenarios in advance. The policy is optimized with Adam and Equation~\ref{Eq:Supervised Learning} for 50 epochs. The test is performed only once.

	\textbf{DAgger}. The behavior policy is set to $\pi_{\theta}$. The $\pi^{ref}$ is recorded during the training process. For every 50,000 frames, the data is collected to aggregate the current dataset. The aggregated dataset is used to train the policy $\pi$ using Equation~\ref{Eq:Supervised Learning}. The optimization includes 20 epochs of training (one iteration). After each iteration we perform tests using $\pi_{\theta}$. We run this process until 1 million frames.

	\textbf{RL}. The RL group uses PPO (\cite{schulman2017proximal}) with GAE (\cite{schulman2015high}), and use collision as its feedback. Every iteration contains 10,000 frames of data and 20 epochs of training. The test is performed after each training epoch. We ran the training process up to 1 million frames.

	\textbf{IARL}. The policy is optimized using Algorithm~\ref{Algorithm:IARL}, the training trajectories include no collision. Both IARL(No Imitation) and IARL(Imitation) are tested. The periods of training and testing is kept equal to the RL group.
    
\subsection{Simulation Results and Discussions}

\begin{figure}
\centering
    \begin{subfigure}{.32\linewidth}
  		\centering
  		\includegraphics[width=\linewidth]{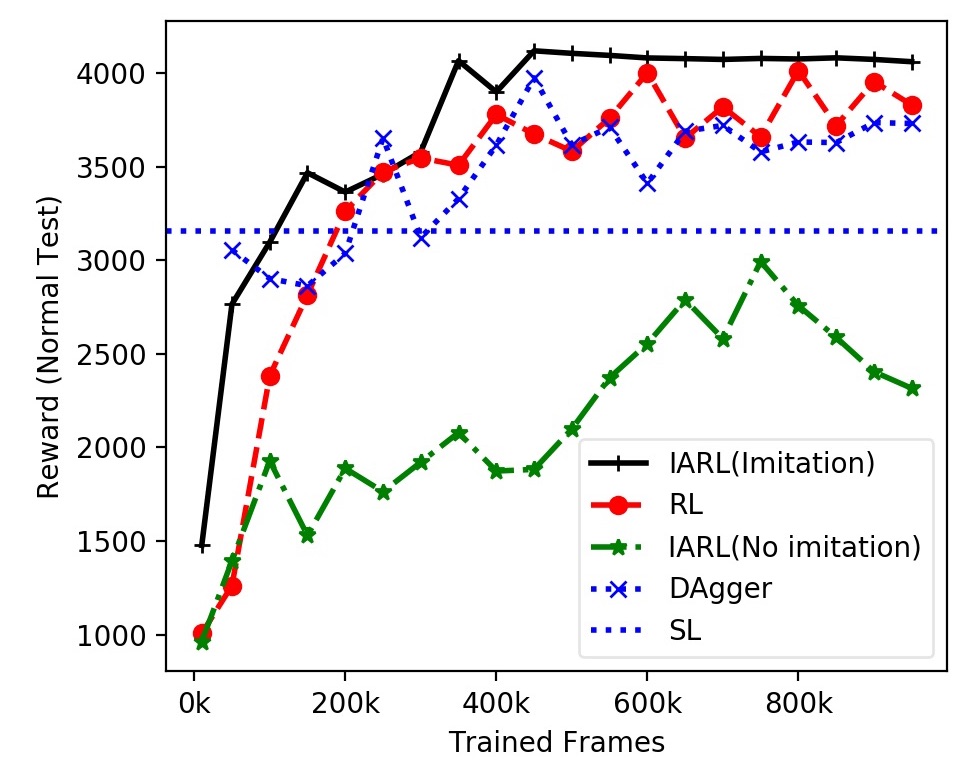}
  		\subcaption{}
  		\label{fig:SimTestResult:sub1}
	\end{subfigure}
	\begin{subfigure}{.32\linewidth}
  		\centering
  		\includegraphics[width=\linewidth]{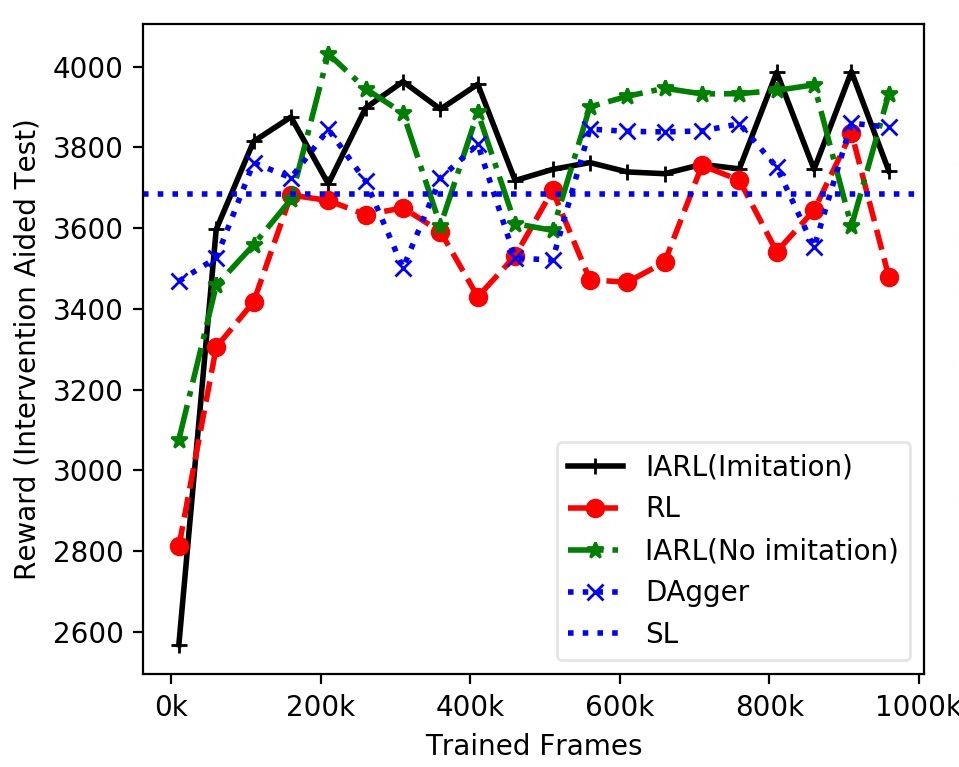}
  		\subcaption{}
  		\label{fig:SimTestResult:sub2}
	\end{subfigure}
	\begin{subfigure}{.32\linewidth}
  		\centering
  		\includegraphics[width=\linewidth]{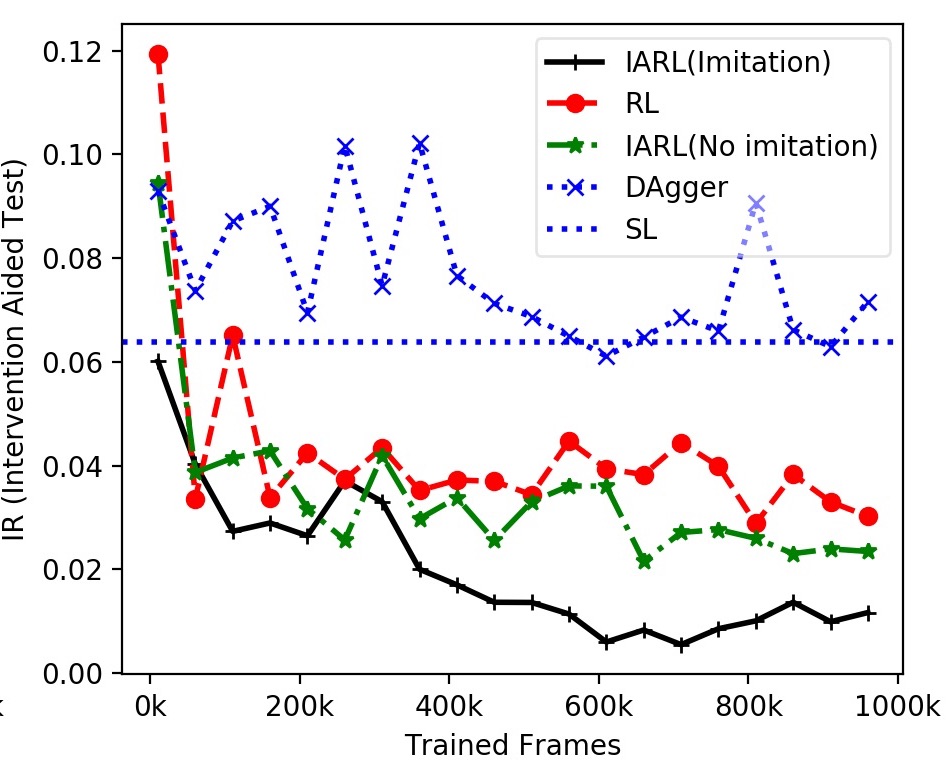}
  		\subcaption{}
  		\label{fig:SimTestResult:sub3}
	\end{subfigure}
	\caption{Comparison of SL, DAgger, RL, IARL(No Imitation) and IARL(Imitation). (a) averaged reward $R_{\text{NT}}$ in the \emph{Normal Test}. (b) averaged reward $R_{\text{IAT}}$ in the \emph{Intervention Aided Test}. (c) averaged \textbf{IR} in the \emph{Intervention Aided Test}.}
    \label{fig:SimTestResult}
    \vspace*{-0.1in}
	\end{figure}

	We compare each of the four methods in both \emph{Normal Test} and \emph{Intervention Aided Test}. The performance against trained frames is plotted in Figure~\ref{fig:SimTestResult}. Several remarks can be made from the results. The IARL(No Imitation) group shows completely different performance with different evaluation methods: In the \emph{Intervention Aided Test} where the reward and the behavior policy of the training and the testing are coherent, it achieves acceptable performance in its rewards and \textbf{IR}; In \emph{Normal Test} where the intervention is removed, the IARL(No Imitation) group collides frequently with obstacles and results in a low performance score. On the other side, the IARL(Imitation) surpasses all the other groups in each test, especially in the \textbf{IR} evaluation.
    
	We count the average collision rate (the collided cases out of 10 test scenarios) over the last 5 evaluations in the \emph{Normal Test}, Tab.~\ref{Tab:Collision} presents the result. The IARL(Imitation) reduces its collision to 0 steadily. While the other groups suffer from occasional collision. Notice that in the \emph{Intervention Aided Test} the value of \textbf{IR} never dropped to 0, as we found that the intervention might overreact in some cases. In other words, the expert intervenes even when the drone could have avoided collision all by itself.

\begin{figure}
  	\centering
  	\includegraphics[width=0.95\linewidth]{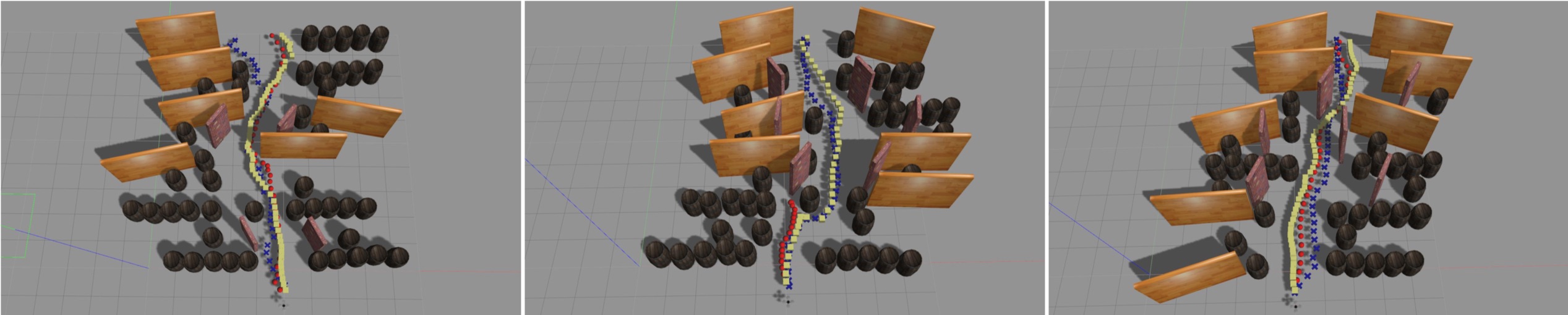}
	\caption{Comparison of trajectories of IARL(Imitation)(Yellow Cubes), RL(Red Spheres), DAgger(Blue Crosses) in three different testing scenarios.}
    \label{fig:Trajectories}
    \vspace*{-.2in}
	\end{figure}

	We also plot the trajectories of RL, IARL(Imitation) and DAgger in Figure~\ref{fig:Trajectories}. A close look at the differences between the IARL and the other two groups shows that the IARL agent is more likely to choose a conservative trajectory. In other words, it is more likely to keep enough distance away from the obstacles, while the RL one pursues more risky trajectories for higher rewards. Besides, we can see that the DAgger and the IARL agent successfully explore a path out of a blind alley(the trajectories in the middle figure), where the RL agent fails to escape(it stops, without any collision). Nevertheless, overstepping the local optimum by exploring around is nearly impossible for reactive controllers that relies on single frame observation only.

\begin{table}
\begin{center} 
  \caption{Average collision in \emph{Normal Test} in different groups}
  \label{Tab:Collision}
  \begin{tabular}{c|c|c|c|c|c}
    \hline
    Method & SL & DAgger & RL & IARL(No Imitation) & IARL(Imitation) \\ 
    \hline
   	Average Collision & 40\% & 24\% & 22.5\% & 42\% & \textbf{0\%} \\
    \hline
  \end{tabular}
\end{center}
\vspace*{-.2in}
\end{table}

\subsection{Experiments in Reality}

\begin{figure}
  	\centering
  	\includegraphics[width=1.0\linewidth]{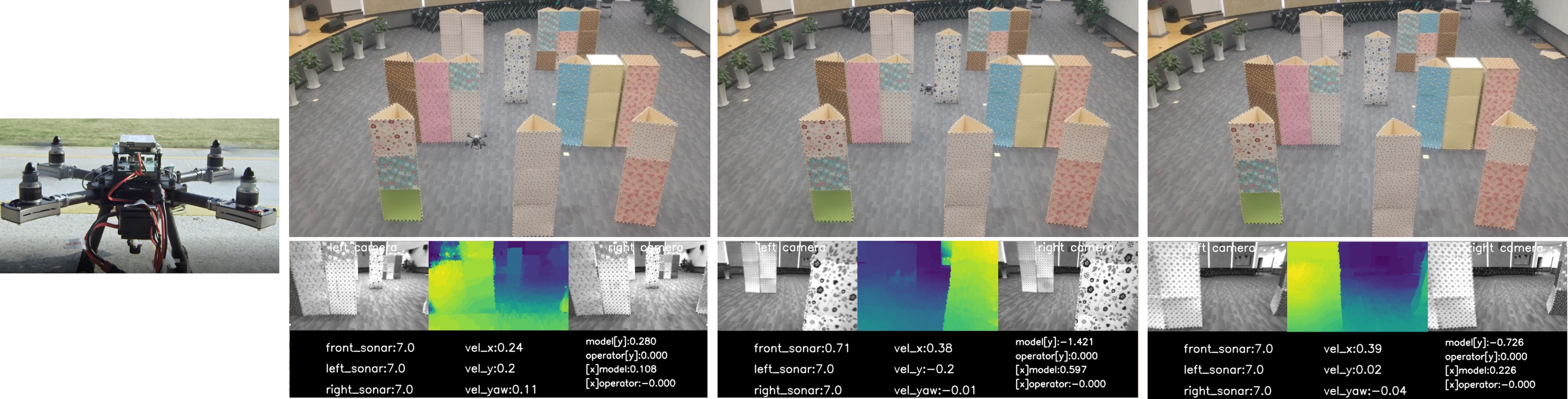}
	\caption{A close-up image of the Drone(left) and three frames during the test. The picture above is captured by external camera, the image below shows the corresponding input and output of the decision model.}
    \label{fig:RealTrajectories}
    \vspace*{-0.05in}
	\end{figure}

	\textbf{Environments.} The drone in real world is a X type quad-rotor helicopter with 38cm size and 1.6kg weight(Figure~\ref{fig:RealTrajectories}). It is equipped with Nvidia Jetson TX2 as the onboard computer, stereo vision, supersonic sensors, IMU and barometer. All computations are completed on board. For simplicity, the environment is composed of two kinds of different obstacles, which are built from foam boards, including trigonometric and square pillars(Figure~\ref{fig:RealTrajectories}). 
    
    \textbf{Start-up Policy.} In order to validate IARL in reality with more acceptable cost, we first collected 100,000 frames of expert manipulation in different cluttered scenarios. We then took advantage of the policy and the value network trained in simulation. We fine-tuned the policy on the labeled data for certain epochs. This model is further used as the start-up policy for IARL training.
    
    \textbf{IARL Training.} During the training process, a human operator keeps watching on the drone and intervening with remote controller when necessary based on his own judgment. The training environments are manually readjusted randomly after several passes. Each iteration(training round) includes 5000 frames of data(around 20 passes) and 20 epochs of training. Extra 2 different scenarios are built and used for test. Similar to the evaluation in the simulations, we apply the \emph{Intervention Aided Test} in the two test scenarios every 20,000 frames, but the \emph{Normal Test} is not available in reality due to the safety issue.

\begin{figure}
  	\centering
    \begin{subfigure}{.3\linewidth}
  		\centering
  		\includegraphics[width=\linewidth]{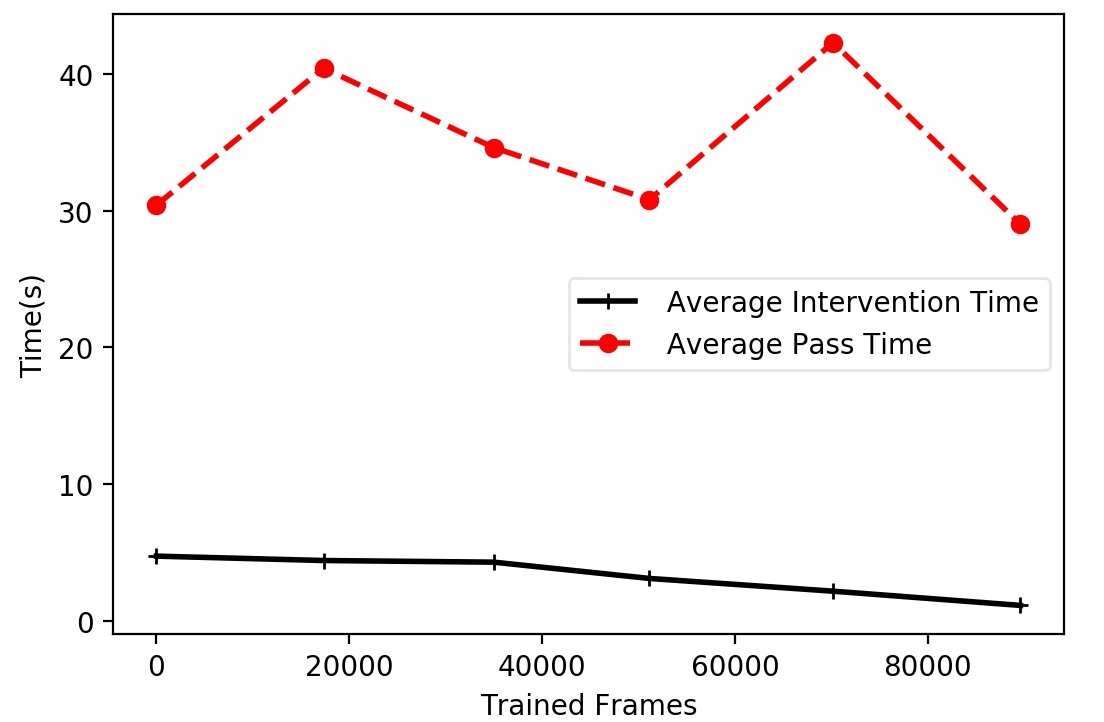}
  		\subcaption{}
  		\label{fig:RealTestResult:sub1}
	\end{subfigure}
    \begin{subfigure}{.3\linewidth}
  		\centering
  		\includegraphics[width=\linewidth]{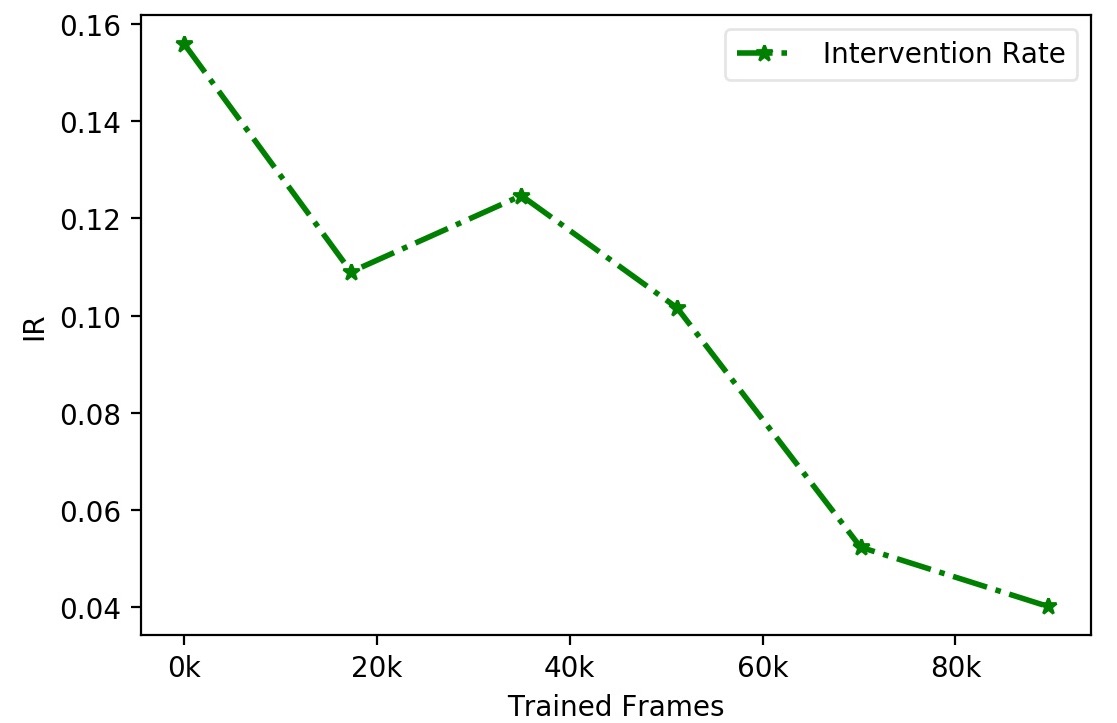}
  		\subcaption{}
  		\label{fig:RealTestResult:sub2}
	\end{subfigure}
	\caption{Different evaluation criteria against the training frames in the IARL training. (a) average pass time and average intervention time. (b) \textbf{IR}.}
    \label{fig:RealResult}
    \vspace*{-0.05in}
	\end{figure}

	\textbf{Results.} The \textbf{IR} in each test round is plotted in Figure~\ref{fig:RealResult}. At the same time, we plot the average pass time(defined as the total time consumed from the start point to the goal) and average intervention time(defined as the length of time that $g_t = 1$). It is worth mentioning that the overall performance is not as good as the simulation(where \textbf{IR} dropped below $1\%$). There can be two reasons: Firstly, the signals acquired from real sensors are much more noisy, especially the depth image; Secondly, there are considerable delay between the observation and the decision making, which is mainly caused by the stereo matching and visual odometer. The delay can go up to 0.3s or 0.4s. Even with those disadvantages, we can see that our model learned to substantially reduce the \textbf{IR} and the intervention time.

\section{Conclusion}
	In this paper, we propose the IARL framework that improves the policy by avoiding and imitating the external intervention at the same time. We successfully deploy our method to UAV platform to achieve map-less collision-free navigation in the unstructured environments. The experimental results show that IARL achieves satisfying collision avoidance and low intervention rate, and ensures safety in the training process. By looking into the future of this technique, we anticipate that IARL will serve as an effective policy optimization metric in various high risk real-world scenarios.

\clearpage


\small{
\bibliographystyle{corlabbrvnat}
\bibliography{IARL}  
}

\newpage
\newcommand{\beginsupplement}{%
  \renewcommand{\thetable}{S\arabic{table}}%
  \renewcommand{\thefigure}{S\arabic{figure}}%
  \renewcommand{\thesection}{S\arabic{section}}%
  \renewcommand{\thesubsection}{S\arabic{section}.\arabic{subsection}}%
}
\beginsupplement
\section{Supplementary Materials}

\subsection{The Value Function Approximation}

In IARL, the approximation of value function $V_{\phi}(s)$ is optimized by minimizing the mean square error loss in Equation~\ref{ValueLoss}.

\begin{equation}
\label{ValueLoss}
L^V(\phi) = \mathbb{E}_{\pi^{mix}_{\theta}}[\sum_t (V_{\phi}(s_t) - \hat{V}_t)^2]
\end{equation}

There are different estimation metrics for $\hat{V}_t$. For example, considering the IARL paradigm, $\hat{V}_t=\sum_{\tau \geq t}\gamma^{\tau - t}r^{mix}_{\tau}$ represents the Monte Carlo(MC) estimation, and $\hat{V}_t=\gamma V_{\phi'}(s_{t+1}) + r^{mix}_t$ represents the temporal difference estimation, where $\phi'$ is the parameters of the target network. In this work we use n-step return estimation shown in Equation~\ref{ValueEstimate}.

\begin{equation}
\label{ValueEstimate}
\hat{V}_t=\sum_{t \leq \tau \leq T}\gamma^{\tau - t}r^{mix}_{\tau} + \gamma^{T - t + 1} V(s_{T+1})
\end{equation}

The difference between Equation~\ref{ValueEstimate} and the MC estimation is that we backup the value function of the last state $s_{T+1}$, which is to avoid the possible bias brought by the finite horizon.

\subsection{Extending IARL(Imitation) to PPO}

The policy loss of PPO based IARL is written as Equation~\ref{IARL-PPO}. $\theta'$ denotes the parameter $\theta$
from the last iteration. $\beta$ is dynamically controlled by another hyper-parameter $d_{targ}$. For more details please refer to \cite{schulman2017proximal}.

\begin{equation}
\label{IARL-PPO}
\begin{split}
&L^{IARL-IL-PPO} = - \mathbb{E}_{\pi^{mix}_{\theta'}}[\sum_{g_t = 0} \Psi_t \frac{\pi_{\theta}(a_t|s_t)}{\pi_{\theta'}(a_t|s_t)} - \alpha \sum_{g_t = 1} (a^{ref}_t - \mu_{\theta}(s_t))^2 \\
&\qquad - \beta \sum_{g_t = 0} KL(\pi_{\theta'}(\cdot|s_t) || \pi_{\theta}(\cdot|s_t))]]
\end{split}
\end{equation}

\subsection{Features of Observation}

At each time step, the policy and value networks take 25 frames of observation and encode the observations using LSTM unit. Each frame of observation is composed of the features listed in Table.~\ref{Tab:Features}

\begin{table}[h!]
\begin{center} 
  \caption{Features in each frame of observation}
  \label{Tab:Features}
  \begin{tabular}{m{6cm}<{\centering}|m{3cm}<{\centering}}
    \hline
    Description & Dimension \\ 
    \hline
    Depth Images, acquired from stereo matching & $45 \times 30$ \\
    \hline
    Velocities in UAV body coordinate system, acquired from Visual Odometer, discretized independently in x and y directions & $20 \times 2$ \\ 
    \hline
    Distance reported by supersonic sensors, discretized & $20 \times 3$ \\ 
    \hline
    Angle of direction to the goal, discretized & $20$\\ 
    \hline
  \end{tabular}
\end{center}
\end{table}

\subsection{Details of The Expert Simulation}

Simulating experts in Gazebo 7 includes $\pi^{ref}$ and $\mathcal{G}$, which represent the expert's policy and intervention probability respectively. 

$\pi^{ref}$: For each new scenario, we discretized the map into $300 \times 300$ grids. We restrict that the UAV moves towards the neighboring 8 grids. The cost of moving to the neighboring grid is defined as the distance to the center of the grid. We then mask all the grids that intersect with the obstacles, and ran Dijkstra's algorithm on the discretized map to acquire the cost to the goal for each grid. During each time step of the training or testing process, starting from the current position, we plan a trajectory by setting magnitude of velocity to 0.20m/s and following the direction to the grid of lowest cost. We used the next three planned positions on the current planned trajectory as the target. We defined the loss as the mean square error between the predicted positions and the target positions. We applied the Model Predictive Control to give $\mu^{ref}$. 

$\mathcal{G}$: We collect a small amount of data in advance to build the $\mathcal{G}$ simulator. Driving the UAV with a RL pre-trained policy(but not yet converged, i.e. an imperfect policy), we required the experts to keep UAV safe by intervening the UAV when necessary using keyboard in Gazebo7 simulation. We collect 50,000 frames of data. Each sample has the following features

\begin{itemize}
\item The roll and pitch angle of the UAV
\item The velocity of the UAV in its body coordinate system
\item Top-10 nearest obstacles, represented by its obstacle type, and the position of the point that is closest to the UAV. The position is calculated in the UAV body coordinate system.
\end{itemize}

Each sample is labeled with $g_t$. We train a Gradient Boosting Decision Trees(GBDT) model with cross-entropy loss, which gives the $\mathcal{G}$.

\subsection{Hyper-parameters}

\begin{table}[h!]
\begin{center} 
  \caption{List of the hyper-parameters in RL and IARL}
  \label{Tab:Hyperparameters}
  \begin{tabular}{m{4cm}<{\centering}|m{2.5cm}<{\centering}|m{6.5cm}<{\centering}}
    \hline
    Hyper-parameter & Value & Description \\ 
    \hline
   	$\gamma$ & 0.96 & Discount factor in RL, IARL \\
   	\hline
   	$\lambda$ & 0.98 & Exponential decay weight in generalized advantage estimation(GAE, \cite{schulman2015high}) in RL and IARL\\
   	\hline
   	Learning rate(policy) & 1.0e-4 & Adam optimizer \\
   	\hline
   	Learning rate(value) & 1.0e-3 & Adam optimizer \\
   	\hline
   	Mini-batch size(value and policy) & 32 & Mini-batch size in RL, IARL and SL \\
   	\hline
   	Iteration size & 1.0e-3 & Collect this many data before training in each iteration in RL and IARL\\
   	\hline
   	Action repeat & 4 & Repeat each action selected by the agent this many times in RL and IARL \\
   	\hline
   	$d_{targ}$ & 0.003 & $d_{targ}$ in PPO(\cite{schulman2017proximal}), for RL and IARL \\
   	\hline
   	$\alpha$ & 2.0 & The ratio of imitation loss and policy loss in IARL \\
   	\hline
   	$\sigma^2$ & 1.8 & The initial exploration noise in policy $\pi_{\theta}$, $\sigma$ is optimized together with $\theta$ during the training process in RL and IARL. \\ 
    \hline
  \end{tabular}
\end{center}
\end{table}

\end{document}